# Multilinear Principal Component Analysis Network for Tensor Object Classification

Rui Zeng, Jiasong Wu, *Member*, *IEEE*, Zhuhong Shao, Lotfi Senhadji, *Senior Member*, *IEEE*, and Huazhong Shu, *Senior Member*, *IEEE*

*Abstract*—The recently proposed principal component analysis network (PCANet) has been proved high performance for visual content classification. In this letter, we develop a tensorial extension of PCANet, namely, multilinear principal analysis component network (MPCANet), for tensor object classification. Compared to PCANet, the proposed MPCANet uses the spatial structure and the relationship between each dimension of tensor objects much more efficiently. Experiments were conducted on different visual content datasets including UCF sports action video sequences database and UCF11 database. The experimental results have revealed that the proposed MPCANet achieves higher classification accuracy than PCANet for tensor object classification.

## I. Introduction

A MAJOR difficulty of object description in visual world comes from the large amount of intra-class variability, arising from illumination, rotation, scaling or more complex deformation. Over the last few years, learning multiple levels representation from visual content by convolutional architecture in deep learning has received much attention. The convolutional architecture imitates the structure of visual system in brain and was shown the ability to learn more robust representation [1]. A convolutional neural network (CNN) [2] generally consists of multiple trainable stages stacked on the top of each other, following a supervised classifier. Each stage of CNN is organized in two layers: convolution layer and pooling layer.

Recently, Chan et al. [3] proposed principal components analysis network (PCANet), which is a convolutional architecture and uses the most basic and simple operations (PCA + binary hashing + block-wise histograms) to emulate the processing layers of CNN. Their method achieves the state-of-the-art performance for most image classification tasks. However, multidimensional patches taken from visual content in PCANet, are simply converted into vector to learn the dictionary in convolutional layer. The vector representation of these patches destroys the spatial structure and the relationship for each dimension in visual content. Moreover, it may also suffer from the so-called curse of dimensionality [4]. Many work [5-7] showed that multidimensional visual content is more suitable to be represented by a tensor object. In the past year, the tensorial extension of deep learning has received great researchers' interest. For example, Hutchinson et al. [8] proposed tensor deep stacking network, which is a tensor extended version of traditional deep neural network (DNN), and applied it for MNIST handwriting image recognition, phone classification and recognition. Yu et al. [9] successfully applied tensor theory to DNN, i.e. deep tensor neural network, which outperforms DNN in large vocabulary speech recognition. To the best of our knowledge, the similar tensorial extension on convolutional architecture has not been reported in the literature.

In this letter, we make a tensorial extension on PCANet and propose multilinear principal components analysis network (MPCANet), whose layers use tensor-to-tensor projection. The basic idea of MPCANet comes from that multilinear principal components analysis (MPCA) [5] outperforms PCA in tensorial dimensionality reduction. We empirically compare the proposed MPCANet with the PCANet on UCF sport action video database [10] and UCF 11 database [11]. The experimental results demonstrate that the MPCANet generally outperforms PCANet in tensor objects classification.

## II. Review of MPCA

MPCA is a tensorial extension of PCA and can capture much more original tensorial input variation than PCA. In this section, we briefly review MPCA [5].

An $N$th-order tensor object is denoted as $X \in \mathbb{R}^{I_1 \times I_2 \times \cdots \times I_N}$. It is represented by $N$ indices $i_n$, $n = 1, 2, \ldots, N$, and each $i_n$ addresses the $n$-mode of $X$. The $n$-mode tensor product of $X$ by a matrix $\mathbf{U} \in \mathbb{R}^{J_n \times I_n}$ is defined as:

This work was supported by the National Basic Research Program of China under Grant 2011CB707904, by the NSFC under Grants 61201344, 61271312, 11301074, and by the SRFDP under Grants 201110092110023 and 20120092120036, the Project-sponsored by SRF for ROCS, SEM, and by Natural Science Foundation of Jiangsu Province under Grant BK2012329 and by Qing Lan Project. This work was supported by INSERM postdoctoral fellowship.

R. Zeng, Z. Shao, and H. Shu are with the LIST, Key Laboratory of Computer Network and Information Integration (Southeast University), Ministry of Education, 210096 Nanjing, China, and also with the Centre de Recherche en Information Biomédicale Sino-Français (CRIBs), 210096 Nanjing, China (e-mail: {zengrui, 230119289, shu.list}@seu.edu.cn).

J. Wu is with the LIST, Key Laboratory of Computer Network and Information Integration (Southeast University), Ministry of Education, 210096 Nanjing, China, INSERM U 1099, 35000 Rennes, France, the Centre de Recherche en Information Biomédicale Sino-Français (CRIBs), 210096 Nanjing, China, (e-mail: jswu@seu.edu.cn).

L. Senhadji is with INSERM U1099, 35000 Rennes, France, with the Laboratoire Traitement du Signal et de l'Image (LTSI), Université de Rennes 1, 35000 Rennes, France, and with the Centre de Recherche en Information Biomédicale Sino–Français (CRIBs), 35000 Rennes, France (e-mail: lotfi.senhadji@univ-rennes1.fr).

$$(X \times_n \mathbf{U})_{(i_1,\ldots,i_{n-1},j_n,i_{n+1},\ldots,i_N)} = \sum_{i_n} X_{(i_1,\ldots,i_N)} \cdot \mathbf{U}_{(j_n,i_n)}. \quad (1)$$

The objective of MPCA is the determination of $N$ projection matrices $\{\mathbf{v}^{(n)^T}\}_{n=1}^N$ to map the tensor set $\{X_m \in \mathbb{R}^{I_1 \times I_2 \times \cdots \times I_n}\}_{m=1}^M$ into $\{Y_m \in \mathbb{R}^{P_1 \times P_2 \times \cdots \times P_n}\}_{m=1}^M$, which satisfy the following rules:

$$X_m = Y_m \times_1 \mathbf{v}^{(1)^T} \times_2 \mathbf{v}^{(2)^T} \times \cdots \times_N \mathbf{v}^{(N)^T},$$
$$\{\mathbf{v}^{(n)} \in \mathbb{R}^{I_n \times P_n}\}_{n=1}^N = \arg\max_{\mathbf{v}^{(1)},\mathbf{v}^{(2)},\ldots,\mathbf{v}^{(N)}} \Psi_Y, \quad (2)$$

where $\Psi_Y = \sum_{m=1}^M \|Y_m - \overline{Y}\|_F^2$, $\overline{Y}$ denotes the mean tensor calculated as $\overline{Y} = (1/M)\sum_{m=1}^M Y_m$.

MPCA is a projection of a high-dimensional tensor to a low-dimensional one of the same order. In practical use, the small volume tensor is then spread into a vector $Z \in \mathbb{R}^{p_1 p_2 \cdots p_N}$, whose elements are arranged according to the variance it's keeping [5]. Fig. 1 provides a visual illustration of conventional MPCA. Note that MPCA degrades to two-dimensional (2-D) PCA [12] and PCA when dealing with matrix (second-order tensor) and vector (first-order tensor), respectively.

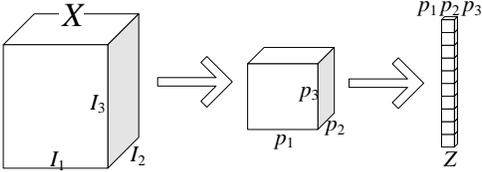

Fig. 1. Visual illustration of MPCA flow diagram

## III. THE ARCHITECTURE OF MPCANET

The architecture of proposed MPCANet is summarized in Fig. 2. In this section, the classification of third-order tensor object via MPCANet is analyzed for simplicity.

### A. Learning projection dictionary

Assume that we have $M$ third-order tensor objects $\{X_m \in \mathbb{R}^{I_1 \times I_2 \times I_3}\}$ for training and corresponding labels. The patch size of tensor objects is set to be $k_1 \times k_2 \times k_3$. We collect all $I_1 \times I_2 \times I_3$ tensor patches by around each element of the $m$th tensor object. We note these tensor patches set by $\{\mathbf{t}_{m,q} \in \mathbb{R}^{k_1 \times k_2 \times k_3}\}_{q=1}^{I_1 \times I_2 \times I_3}$. Repeating the above process for every tensor object, we can get all tensor patches $\mathbf{t} = \{\mathbf{t}_m\}_{m=1}^M$ for training convolutional dictionaries. Then MPCA is applied to $\mathbf{t}$ to get three projection matrices $\{\mathbf{v}^{(n)} \in \mathbb{R}^{k_n \times P_n}\}_{n=1}^3$. $\mathbf{v}^{(n)}$ are used for encoding the tensor object in the projected encoder layer.

### B. Projected encoder layer

The patch set $\mathbf{t}_m$ is projected to $\{\mathbf{s}_{m,q} \in R^{p_1 \times p_2 \times p_3}\}_{q=1}^{I_1 \times I_2 \times I_3}$ by using the above projection dictionaries $\mathbf{v}^{(n)}$. Each element of $\mathbf{s}_m$ is converted into vector according to its variance and then formed a vector set $\{\mathbf{z}_{m,q} \in \mathbb{R}^{p_1 p_2 p_3}\}_{q=1}^{I_1 \times I_2 \times I_3}$. Assuming that the number of encoders is $L$. For each vector in $\mathbf{z}_{m,q}$, we pick the first $L$ elements and construct a matrix for putting these vector together.

$$\overline{\mathbf{z}}_m = \begin{bmatrix} \mathbf{z}_{m,1(1)} & \mathbf{z}_{m,1(2)} & \cdots & \mathbf{z}_{m,1(L)} \\ \mathbf{z}_{m,2(1)} & \mathbf{z}_{m,2(2)} & \cdots & \mathbf{z}_{m,2(L)} \\ \vdots & \vdots & \ddots & \vdots \\ \mathbf{z}_{m,I_1 \times I_2 \times I_3(1)} & \mathbf{z}_{m,I_1 \times I_2 \times I_3(2)} & \cdots & \mathbf{z}_{m,I_1 \times I_2 \times I_3(L)} \end{bmatrix}. \quad (3)$$

All columns of matrix $\overline{\mathbf{z}}_m$ are converted into a new tensorial feature set $\{F_m^l \in \mathbb{R}^{I_1 \times I_2 \times I_3}\}_{l=1}^L$, where $F_m^l$ is the $l$th tensorial feature of $X_m$.

### C. Tensor feature pooling

First, to make tensorial feature more robust, we binarize each tensorial feature by using Heaviside step function $H(\cdot)$, whose values are one for positive entries and zero otherwise. The binarized tensorial feature is denoted as $\overline{F}_m^l$. Since each tensorial feature keeps different variances of original data, $\overline{F}_m^l$ should be weighted to form a new single tensor feature as follows:

$$W_m = \sum_{l=1}^L 2^{l-1} \overline{F}_m^l. \quad (4)$$

Note that each element of tensorial feature $W_m$ is an integer in the range $[0, 2^L-1]$.

Next, a spatial pooling operation is applied to $W_m$. The cubic feature is divided into $B$ boxes. We compute histogram of the decimal values for each box and denote it as $\text{hist}(\text{Box})_b$. After the above pooling process, we simply concatenate all the histograms of $B$ boxes into one vector, i.e.,

$$f_m = [\text{hist}(\text{Box})_1, \text{hist}(\text{Box})_2, \ldots, \text{hist}(\text{Box})_B] \in \mathbb{R}^{2^L B}. \quad (5)$$

Boxes can be either overlapping or non-overlapping in MPCANet, depending on applications [3].

### D. A multi-stage MPCANet

Now we are ready to describe the multi-stage MPCANet. As depicted in Fig. 2. The two-stage MPCANet (MPCANet-2) contains two projected encoder layers (C1 and C2) and a pooling layer. Assuming that the encoder number in C1 and C2 are $L_1$ and $L_2$, respectively. The projection dictionary $\mathbf{v}^1$ in C1 layer is obtained from tensor objects set $\{X_m \in \mathbb{R}^{I_1 \times I_2 \times I_3}\}_{m=1}^M$. We use $\mathbf{v}^1$ to map $X_m$ to a new tensorial feature set $\{F_m^l \in \mathbb{R}^{I_1 \times I_2 \times I_3}\}_{l=1}^{L_1}$. In C2 layer, we get projection dictionary $\mathbf{v}^2$ by using all tensorial features of all tensor objects. Then C2 layer projects $F_m^l$ to new tensorial feature set $\{G_m^{lh} \in \mathbb{R}^{I_1 \times I_2 \times I_3}\}_{h=1}^{L_2}$, where $G_m^{lh}$ denotes the $h$th tensorial feature of $F_m^l$. We binarize $G_m^{lh}$ and then weight them to get $W_m$, shown in (4). Pooling operation is applied to these $L_1 \times L_2$ tensorial features, and then we get final feature vector $f_m$ for $X_m$. One or more additional layers can be stacked like C1-C2-C3… if a deeper architecture is found to be beneficial.

## IV. Experimental Results

### A. Performance in UCF sports action database

The experiments are carried out on UCF sports action video dataset. It contains approximately 200 sport action videos at resolution of $720 \times 480$, included in 9 classes, which are typically featured on broadcast television channels such as the BBC and ESPN.

Before feeding the UCF samples to MPCANet, the tensorial input needs to be normalized to the same dimension in each mode. Every video is converted from color to gray. The primary individual, who doing sport, in each frame, is cropped with dimension $400 \times 250$, which is then resized to $80 \times 50$ pixels in order to reduce the computational complexity. The normalized time-mode dimension is chosen to be 20, which keeps a complete action as much as possible. Thus, each video sample has a canonical representation of $I_1 \times I_2 \times I_3 = 80 \times 50 \times 20$. We randomly pick up half videos in each class as training video, and the others as testing video.

Since the MPCA is the tensorial extension of the conventional PCA, we use consistent parameters with that of PCANet reported in [3]. To keep computational simplicity, we do not move patch in the direction of third-mode of the tensor object. The patch size is set to be $3 \times 3 \times 20$, $5 \times 5 \times 20$, and $7 \times 7 \times 20$. Thus, the tensorial feature in C2 is the second-order tensor (matrix). There is no effect on other layers. We always set $L_1 = L_2 = 8$ for all networks. In order to learn correctly projection dictionaries, 97% energy of patches is kept [5] in each projection dictionary layer of MPCANet. The box sizes in pooling layer for MPCANet and PCANet are set to be the multiple of $8 \times 5$. To offer some degree of translation invariance in tensor objects, the overlapping ratio of box is set to 50%.

We also compare the proposed MPCANet with conventional MPCA+LDA [5]. The input of MPCA+LDA is the entity of tensor objects. The dimensions of MPCA feature vector, which is then putted into LDA, vary from 10 to 100.

MPCANet-2 has two different forms: MPCANet-2-Cuboid and MPCANet-2-Vector. The differences between them are the shape of patch in C2 layer. For the former one, we treat tensor patch as a tensor object to learn projection dictionary. For the latter one, tensor patches are converted into a vector to learn corresponding MPCA projection dictionary.

The recognition rates of above networks and MPCA+LDA averaged over 5 different random splits are shown in Fig. 3. The best performances of MPCANet, PCANet, and MPCA+LDA are listed in Table I.

It can be observed that both MPCANet-1 and PCANet-1 outperform MPCA+LDA. The reason is that convolution architecture imitates visual system in human brain, it can provide more robust features for visual content [1].

PCANet-1 achieves the best performance among one-stage networks, but the improvement from PCANet-1 to PCANet-2 is not larger as that of MPCANet. One can see that MPCANet-2-vector achieves the best recognition result in two-stage networks. MPCANet-2-vector utilizes the spatial structure and the relationship between each dimension in projection dictionary layer. This operation retains more infor-mation of tensor objects than PCANet-2. On the whole, MPCANet-2-cubic performs the worst in tensor objects classification among six networks. Why the classification accuracy difference between MPCANet-2-cubic and MPCANet-2-vector is so large? Because in the experiment, we have already use MPCA to reduce the redundancy of the third mode (i.e., time dimension) in tensor object in C1 layer. It means that, for C2 layer, the patches do not have great size. So, it is better to treat these tensor patches as vector to learn the projection dictionary.

TABLE I
THE BEST PERFORMANCE OF MPCANET, PCANET [3], AND MPCA+LDA [5] ON UCF SPORTS ACTION DATABASE

| Methods | Accuracy (%) |
|---|---|
| MPCANet-1 | 63.01 |
| MPCANet-2-vector | **73.93** |
| MPCANet-2-cubic | 52.05 |
| PCANet-1 [3] | 67.12 |
| PCANet-2 [3] | 68.49 |
| MPCA+LDA [5] | 42.34 |

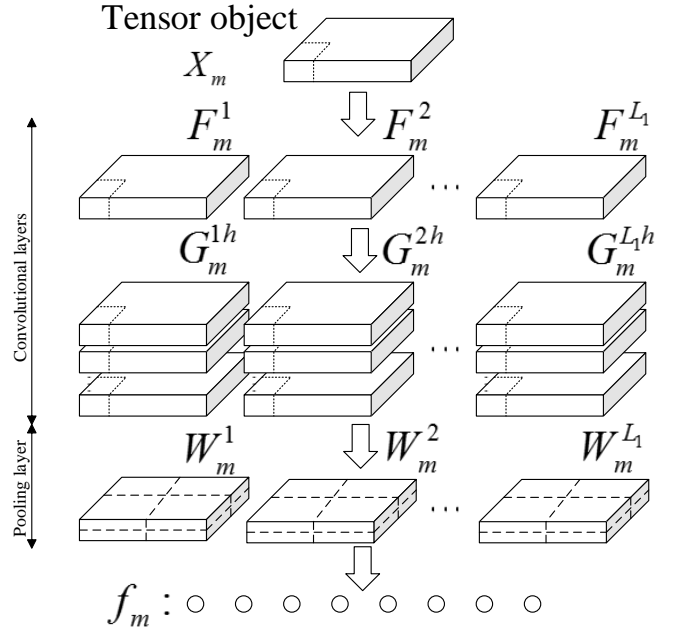

Fig. 2. Architecture of two-stage MPCANet (MPCANet-2)

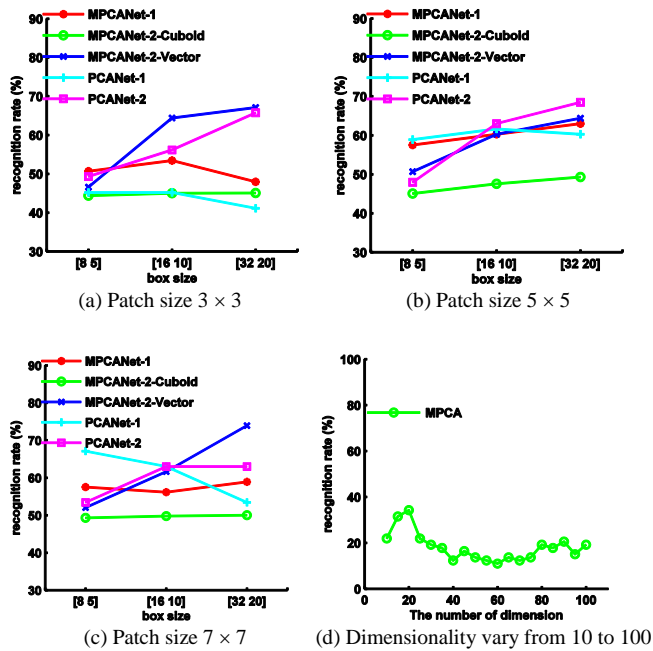

Fig. 3. Recognition rates of MPCANet and PCANet in different patch size $k_1 \times k_2$. (a) $3 \times 3$. (b) $5 \times 5$. (c) $7 \times 7$. (d) is the performance of MPCA+LDA. The number of features varies from 10 to 100.

### B. Testing on UCF11 database

Subsequently, we test the performance of proposed MPCANet on UCF11 for tensor objects classification. UCF11 contains 11 action categories: basketball shooting, biking, diving, golf swinging, horseback riding, soccer juggling, swinging, tennis swinging, trampoline jumping, volleyball spiking, and walking with a dog. They are all $240 \times 320$ pixels action videos and manually collected from YouTube. For each category, the videos are grouped into 25 groups with more than 4 action clips in it. We only choose the top-ten group in each category to test the performance of MPCANet, PCANet [3], and MPCA+LDA [5]. The total number of experimental videos is 642. For each group, half videos are randomly selected for training and others for testing. The UCF11 videos have some variation in frame. For time-mode larger than 20, we only choose the first twenty time modes. For a few videos, whose frames are less than 20, we just copy the last frame to fill them. Every videos are resized to $48 \times 64$ to be easily calculated.

The patch size and overlapping ratio are the same as above experiment shown in UCF sports action database. Three box size sizes $6 \times 8$, $12 \times 16$, $24 \times 32$ are considered here. For MPCA+LDA, the dimensions of feature, which are extracted from MPCA, are changed from 10 to 100. The best performance of MPCANet, PCANet, and MPCA+LDA is listed in Table II.

## V. CONCLUSIONS

In this paper, we proposed and implemented a novel deep learning architecture, that is, MPCANet, which involves tensor interactions among stages. MPCANet is composed of projec-tion dictionaries, projected encoder, and pooling layer. We have described an approach to map the tensor objects to a corresponding feature vector. We have evaluated the performance of the MPCANet on UCF sports action dataset and UCF11. The experimental results demonstrate that MPCANet outperforms conventional PCANet and MPCA+LDA in tensor objects classification. It provides us the inspiration to dealing with tensor objects for other convolutional deep architectures.

TABLE II
THE BEST PERFORMANCE OF MPCANET, PCANET [3], AND MPCA+LDA [5] ON UCF11

| Methods | Accuracy (%) |
| --- | --- |
| MPCANet-1 | 59.53 |
| MPCANet-2-Vector | **79.26** |
| MPCANet-2-Cuboid | 57.12 |
| PCANet-1 [3] | 58.68 |
| PCANet-2 [3] | 76.92 |
| MPCA+LDA [5] | 45.15 |